\pdfoutput=1
\documentclass[11pt]{article}
\usepackage{acl}
\usepackage{times}
\usepackage{latexsym}
\usepackage[T1]{fontenc}
\usepackage[utf8]{inputenc}
\usepackage{microtype}

\usepackage{subfigure}

\usepackage{times}
\usepackage[utf8]{inputenc}
\usepackage{bm}

\usepackage{amsmath}
\usepackage[ruled,linesnumbered,vlined]{algorithm2e}
\usepackage{algpseudocode}

\newcommand{\zstroke}{%
  \text{\ooalign{\hidewidth -\kern-.3em-\hidewidth\cr$z$\cr}}%
}

\usepackage{url}
\usepackage{hyperref}
\hypersetup
{
    colorlinks=true,
    citecolor=navyblue,
    linkcolor=navyblue,
    filecolor=navyblue,
    urlcolor=navyblue,
}

\usepackage{tabularx}
\usepackage{booktabs}
\usepackage{makecell}
\usepackage{diagbox}
\usepackage{multirow}
\usepackage{caption}
\newcolumntype{s}{>{\hsize=.5\hsize}X}

\usepackage{natbib}

\newcommand{\fn}[1]{\footnote{#1}}
\usepackage{lipsum}

\usepackage{xcolor}
\usepackage{todonotes} 

\usepackage{soul}
\renewcommand{\hl}[2][blue]{\textcolor{#1}{#2}}
\newcommand{\mycomment}[1]{\textit{#1}}
\renewcommand{\hl}[1]{#1}
\renewcommand{\todo}[1]{}
\renewcommand{\mycomment}[1]{}

\usepackage{CJKutf8}

\usepackage{comment}

\usepackage{verbatim}

\usepackage{colortbl}
\definecolor{cR}{RGB}{255,204,204}
\definecolor{cG}{RGB}{204,255,204}
\definecolor{cB}{RGB}{204,204,255}
\definecolor{cK}{RGB}{221,221,221}
\definecolor{navyblue}{rgb}{0.0,0.0,0.5}
\newcommand{\bgR}[1]{\cellcolor{cR}#1}

\newcommand{\bgB}[1]{\cellcolor{cB}#1}
\newcommand{\bgK}[1]{\cellcolor{cK}#1}

\newcommand{\Sec}[1]{Section~\ref{sec:#1}}
\newcommand{\Fig}[1]{Figure~\ref{fig:#1}}
\newcommand{\Tab}[1]{Table~\ref{tab:#1}}
\newcommand{\Algo}[1]{Algorithm~\ref{algo:#1}}

\newcommand{\ra}{{$\rightarrow$}}
\newcommand{\task}[2]{\mbox{#1{\ra}#2}}

\newcommand{\argmax}{\mathop{\rm arg~max}\limits}

\makeatletter
\def\adl@drawiv#1#2#3{%
        \hskip.5\tabcolsep
        \xleaders#3{#2.5\@tempdimb #1{1}#2.5\@tempdimb}%
                #2\z@ plus1fil minus1fil\relax
        \hskip.5\tabcolsep}
\newcommand{\cdashlinelr}[1]{%
  \noalign{\vskip\aboverulesep
           \global\let\@dashdrawstore\adl@draw
           \global\let\adl@draw\adl@drawiv}
  \cdashline{#1}
  \noalign{\global\let\adl@draw\@dashdrawstore
           \vskip\belowrulesep}}
\makeatother

\title{Bilingual Corpus Mining and Multistage Fine-Tuning for Improving Machine Translation of Lecture Transcripts}

\author{
Haiyue Song$^{1,2}$ \quad Raj Dabre$^{1}$ \quad Chenhui Chu$^{2}$ \quad Atsushi Fujita$^{1}$ \quad Sadao Kurohashi$^{3}$ \\
$^{1}$NICT / Kyoto, Japan \quad $^{2}$Kyoto University / Kyoto, Japan \quad $^{3}$NII / Tokyo, Japan \\
{\tt \{haiyue\_song, raj.dabre, atsushi.fujita\}@nict.go.jp} \\
{\tt chu@i.kyoto-u.ac.jp} \quad {\tt kuro@i.kyoto-u.ac.jp}
}
\begin{document}
\maketitle
%

\begin{abstract}
Lecture transcript translation helps learners understand online courses, however, building a high-quality lecture machine translation system lacks publicly available parallel corpora. 
To address this, we examine a framework for parallel corpus mining, which provides a quick and effective way to mine a parallel corpus from publicly available lectures on Coursera.
To create the parallel corpora, we propose a dynamic programming based sentence alignment algorithm which leverages the cosine similarity of machine-translated sentences.
The sentence alignment $F_1$ score reaches 96\%, which is higher than using the BERTScore, LASER, or sentBERT methods.
For both English--Japanese and English--Chinese lecture translations, we extracted parallel corpora of approximately 50,000 lines and created development and test sets through manual filtering for benchmarking translation performance.
Through machine translation experiments, we show that the mined corpora enhance the quality of lecture transcript translation when used in conjunction with out-of-domain parallel corpora via multistage fine-tuning. 
Furthermore, this study also suggests guidelines for gathering and cleaning corpora, mining parallel sentences, cleaning noise in the mined data, and creating high-quality evaluation splits. For the sake of reproducibility, we have released the corpora as well as the code to create them. The dataset is available at \url{https://github.com/shyyhs/CourseraParallelCorpusMining}.
\end{abstract}


\section{Introduction}
\label{sec:introduction}

Massive open online courses (MOOCs)\fn{\url{http://mooc.org}} have proliferated, enabling people to attend lectures regardless of their geographical location.
Typically, such lectures are taught by professors from various universities and are made available through video recordings.  
The lectures are usually taught in one particular language, and the video is accompanied by transcripts. The transcripts are typically made by volunteers or an automatic high-accuracy speech recognition system.
To distribute knowledge to a more significant number of people speaking other languages, these transcripts are then translated into other languages.
\begin{figure}[t]
    \centering
    \includegraphics[width=0.65\linewidth]{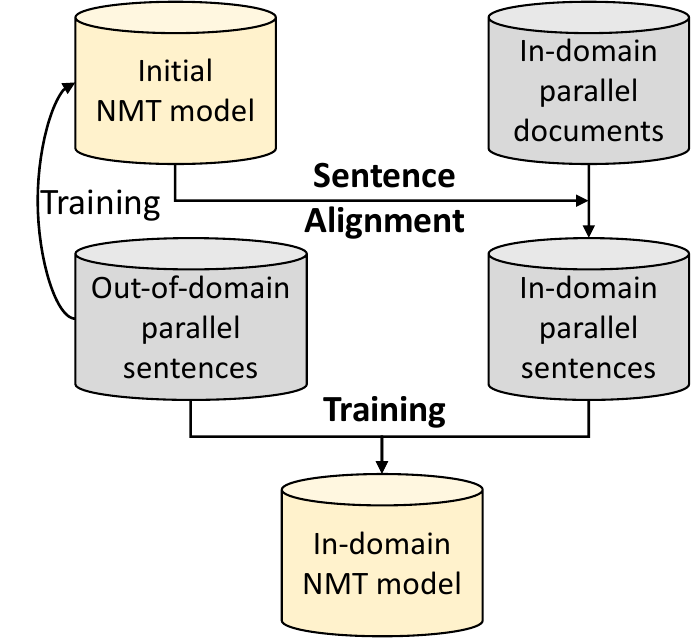}
    \caption{\hl{\textnormal{\textbf{Overview of the corpus creation framework.} In-domain parallel documents are crawled from the Coursera website. The machine translation model trained on out-of-domain data maps data to the same space during sentence alignment.
    }}}
    \label{fig:framework}
\end{figure} 

A high-quality machine translation (MT) system is desirable to translate lecture transcripts efficiently since manual transcript translation is a time-consuming task given that there is an enormous volume of online lectures in different languages.
For lecture transcripts translation, there is sufficient work for lectures in European languages but none for Japanese and Chinese languages.
The TraMOOC project \cite{kordoni2016tramooc} aims to improve the accessibility of online lectures in European languages through MT. They focus on collecting translations of lecture transcripts and constructing MT systems for 11 European languages. 
However, the number of parallel resources involving other languages, such as Chinese and Japanese, is still relatively low compared to European languages. 

Both translating from English to other languages and from other languages to English are urgent and essential.
Given that most lectures are spoken in English, and non-English speakers worldwide greatly outnumber English speakers, translating English transcripts to other languages can help knowledge distribution.
Meanwhile, lectures are taught in languages other than English as well. For instance, several universities in Japan and China offer online courses taught mainly in Japanese and Chinese, respectively. It is also essential to enable non-Japanese and non-Chinese speakers to participate in these courses by translating lecture transcripts into other languages, especially English.

Subtitle\footnote{Lecture transcripts are also known as subtitles, and in this paper, we use these terms interchangeably.} translation falls simultaneously under the spoken language domain and the scientific domain, wherein there are several differences between both domains. Firstly, compared with spoken language domain data such as transcripts of TED talks \citep{cettolo2012wit3}, university lectures are devoted mainly to educational purposes, and there are more terminologies in the lectures. Secondly, compared with scientific domain data in the ASPEC corpus \cite{nakazawa2016aspec}, lecture transcripts contain many spoken features. 
\hl{The subtle difference in the domain may hinder translation quality unless appropriate data reflecting the properties of both domains are used for developing lecture translation systems.} 

Obtaining a parallel corpus is the first step in training the MT system, and applying automatic methods to obtain high-quality parallel data is more desirable than manual methods. Manual methods such as hiring professional translators or employing crowdsourcing platforms are economically infeasible, especially for universities and other non-profit organizations, \hl{because the desired size of the MT dataset is usually large, often ranging from tens of thousands to hundreds of thousands of examples.} 
There are many automatic methods for extracting parallel sentences from roughly parallel documents (also called comparable corpora) \citep{tiedemann2012parallel}.\fn{cf. Comparable corpora, such as Wikipedia, contain pairs of documents comprising the contents on the same topic. However, their parallelism is not necessarily guaranteed, and the corresponding sentences are not necessarily in the same order \citep{Chu:2015:IPS:2856425.2833089}.}
In particular, MT-based approaches are quite desirable because of their simplicity and the possibility of using existing translation models to extract additional parallel data.
However, if there exists no in-domain MT model, using an MT system trained on data from another domain can yield unreliable translations, which can lead to extracted parallel data of low quality.
In addition to using parallel corpora, leveraging in-domain monolingual corpora through back-translation~\citep{sennrich-etal-2016-improving} is also an effective way to improve the performance of the MT system. In this work, we focus on constructing parallel corpora not only because they usually bring larger improvement~\citep{edunov-etal-2018-understanding}, but also because we can build a test set to benchmark the lecture domain MT systems.

We propose a framework shown in \Fig{framework} to create high-quality English to Japanese (Chinese) and Japanese (Chinese) to English MT systems for lecture transcripts translation.
It contains: 1) crawling and cleaning parallel documents\footnote{Parallel documents are also known as comparable documents which are documents containing the same information in different languages.}, 2) extracting in-domain parallel sentences using an initial NMT system trained on out-of-domain corpora and proposed alignment algorithm, and 3) leveraging both the lecture domain corpora and out-of-domain corpora to improve the MT system for lecture transcripts. Our contributions are as follows:

First, we crawl and clean parallel lecture transcript document pairs from Coursera.\fn{\url{https://www.coursera.org}} We download the transcripts in all available languages for each course in Coursera and clean noisy data.

Second, we propose a novel automatic alignment method based on cosine similarities of sentence vector representations by the word2vec model and dynamic programming (DP) algorithms to extract parallel sentences from roughly aligned transcript document pairs. \hl{The proposed alignment algorithm achieved a higher $F_1$ score than methods using the BLEU-score-based method, LASER, or sentBERT. Tested on four manually created document pairs, we show that the proposed method achieved a 96\% $F_1$ score. In contrast, the BLEU-based DP method, BERTScore based DP method, LASER embedding based greedy method, and LASER embedding based DP method achieved 93\%, 89\%, 75\%, and 90\% $F_1$ scores, respectively.} 

Third, with the alignment algorithm, we created two corpora from lectures in Coursera.
The English--Japanese parallel corpus contains 50,543 lines in the train set, 2,068 lines in the test set, and 555 lines in the development set. We do manual filtering for the test and development sets to ensure quality.
In contrast to our previous paper \cite{song-EtAl:2020:LREC3}, we created a high-quality English--Chinese parallel corpus of 40,074 lines with a manually checked test set with 2,009 sentences and a development set with 865 sentences.
We have made the data publicly available so that other researchers can use it to benchmark their MT systems.\fn{\url{https://github.com/shyyhs/CourseraParallelCorpusMining}} 

Finally, we demonstrate the effectiveness of the proposed corpora through MT experiments, showing that the datasets are especially useful when combined with larger out-of-domain corpora using domain adaptation techniques.
Specifically, we explored the potential of multiple datasets for educational lecture translation. Following the curriculum learning paradigm, \hl{we modified the commonly used two-stage fine-tuning method to a multistage fine-tuning strategy.}
Compared with the fine-tuning method \cite{1604.02201}, the multistage fine-tuning model achieved up to 4.1 BLEU score improvements on English and Japanese (Chinese) lecture transcripts MT settings. The multistage method gave the best BLEU scores for all the translation directions.

\section{Related Work}
\subsection{Educational Domain Corpora}


The TraMOOC project \citep{kordoni2016tramooc,kordoni2016enhancing} aims to provide access to multilingual (mainly European languages) transcripts of online courses using MT and provides parallel corpora to train the MT system. The AMARA platform \citep{jansen2014amara} aims at European languages, such as German, Polish, and Russian. However, there are also many Japanese MOOCs such as The Japan MOOC \fn{\url{https://gacco.org}} and Chinese MOOCs such as CNMOOC\fn{\url{https://www.cnmooc.org/home/index.mooc}} that contain numerous courses in Japanese and Chinese. To translate them into English and vice versa, creating a Japanese (Chinese)--English parallel dataset is the first step.

The domain difference is another issue. There are spoken language corpora such as TED talks \citep{cettolo2012wit3}.\fn{\url{https://wit3.fbk.eu/mt.php?release=2017-01-ted-test}} and OpenTranscripts \citep{tiedemann2012parallel}\fn{\url{https://www.opentranscripts.org}}. Moreover, there is ASPEC dataset which contains English--Japanese parallel data from scientific papers. However, the first two are not in the scientific domain and ASPEC is not in the spoken language domain. There are yet no corpora for Japanese (Chinese)--English that falls in both the scientific and spoken language domains.

Therefore, we collected educational domain data from the Coursera website which is a prevalent platform for MOOCs and many lectures have multilingual transcripts. Additionally, they are created by professional and non-professional high-level human translators, ensuring the high-quality of data.

\subsection{Automatic Parallel Sentences Extraction}
Automatic sentence alignment can extract parallel sentences that are orders of magnitude larger than those obtained by manual translation. If the crawled data in the source and target languages is sentence-level, then the neural MT based method~\citep{chen-etal-2020-parallel-sentence} can achieve good performance, pretrained model~\citep{zhang-etal-2020-parallel-corpus} can filter noisy data, and the LASER tool for bitext mining~\citep{chaudhary2019low}\fn{\url{https://github.com/facebookresearch/LASER}} with greedy algorithm also works well. If the crawled data is document-level, MT-based \citep{sennrich2010mt,sennrich2011iterative,liu2018chinese} and similarity-based methods \citep{bouamor2018h2,wang2019target,bert-score} can give more accurate sentence alignments.
The transcripts are crawled from Coursera. They are in the document format where each document contains the transcript of one language from one course. Additionally, lines in the document are in the time order. Therefore, we use this feature and propose our algorithm using dynamic programming for alignment \citep{utsuro1994bilingual}. Previous work Vecalign~\citep{thompson-koehn-2019-vecalign} uses LASER embedding and DP algorithm. 

They rely on the similarity of source and target sentence embeddings that are automatically mapped into the multilingual embedding space.
However, we show that translating data to the same language by an NMT system and utilizing similarities in the monolingual embedding space gives a more robust alignment performance.

\subsection{Domain Adaptation}
Neural machine translation (NMT) systems trained on large parallel corpus provide a high-quality translation~\citep{sutskever2014sequence,bahdanau2014neural,vaswani2017attention, 1604.02201,1706.03872}.
However, we can only construct small-scale datasets for lecture transcript translation due to the scarcity of raw data.

Transfer learning~\citep{chu-etal-2017-empirical, luong2015stanford,sennrich2015improving,1604.02201,raj17, zhang-etal-2019-curriculum, hu-etal-2019-domain-adaptation, zeng-etal-2019-iterative} is widely used to obtain better translation quality for low-resource domains.
The two-stage fine-tuning algorithm is a special case of curriculum learning~\citep{bengio2009curriculum} whose basic idea is to train a model using data ranging from unrelated to related in the learning stages. It is inspired by the human learning process and was first explored in the machine learning field, showing that curriculum learning results in better generalization and faster convergence during training. It has been verified and widely applied to tasks in different fields, including object classification \citep{gong2016multi}, text-to-text machine translation \citep{liu-etal-2020-norm, zhou-etal-2020-uncertainty}, and end-to-end speech translation \citep{wang-etal-2020-curriculum}.

In NMT, we usually use two types of datasets during fine-tuning: larger out-of-domain datasets and smaller in-domain datasets. The model trained on the out-of-domain dataset obtains a general knowledge of translation. Through fine-tuning in-domain data as presented, it also acquires domain-specific knowledge and, thus, performs better in the domain of interest. 

There are more than two datasets in our settings, therefore we propose to modify the two-stage fine-tuning~\citep{1604.02201} and mixed fine-tuning~\cite{raj17} methods to a multistage fine-tuning method which makes the task transition smoother and prevents knowledge forgetting of the previous stages.



\section{Corpus Construction Pipeline}
This section describes the pipeline to construct the lecture transcripts corpus, including 1) crawling and cleaning the parallel lecture transcripts documents from Coursera as described in~\Sec{crawl_clean_docs}, 2) extracting parallel sentences from parallel documents in~\Sec{parallel_sents_extraction}, and 3) creating high-quality evaluation sets manually in~\Sec{evaluation_sets_creation}. 

\subsection{Crawling and Cleaning Parallel Transcripts Documents}
\label{sec:crawl_clean_docs}
We discuss how to obtain and clean the raw parallel lecture transcripts documents in this subsection.

\noindent \textbf{Crawling.} We obtain raw lecture transcripts documents from Coursera by crawling. First, we scrape the list of names of available courses on the Coursera website. 
Then, we use the tool Coursera-dl\fn{\url{https://github.com/coursera-dl/coursera-dl}} to download the transcripts in all available languages for each course in the list. 
Each transcript document contains sentences in the time order, and the multilingual documents for one course are roughly aligned.


\noindent \textbf{Data Cleaning.} We applied the following 5-step procedure to clean the raw data:
\begin{enumerate}
\item \underline{Normalizing Text Encoding}: Text in the documents is converted into UTF-8, and variants of character encoding are normalized to the NFKC\footnote{\hl{\url{https://unicode.org/reports/tr15}}} format.
\item \underline{Detecting Language Mismatch}: 
Some transcript document is in a different language than mentioned in its filename. Thus, it is necessary to detect and exclude such mismatches. We use handwritten rules for the target languages English, Japanese, and Chinese considering their special character types. The automatic language detection tool langdetect\fn{\url{https://github.com/Mimino666/langdetect}} \citep{raffel2019exploring} is also an option.
\item \underline{Splitting Lines into Sentences}: Some lines contain more than one sentences. We use punctuation marks as the clue to segment such lines into multiple sentences. We discard files containing no punctuation marks because there is no reliable way to deal with them.
\item \underline{Removing Meta Tokens}: Tokens indicating meta-information, such as ``[Music]'' and ``$<<$,'' are removed to reduce noise.
\item \underline{Eliminating Imbalanced Document Pairs}: Some document pairs are imbalanced in terms of size: one side has twice or more sentences than the other. These document pairs are usually noisy and we eliminate them.
\end{enumerate}

\subsection{Parallel Sentences Extraction}
\begin{figure}[t]
    \centering
    \includegraphics[width=0.99\linewidth]{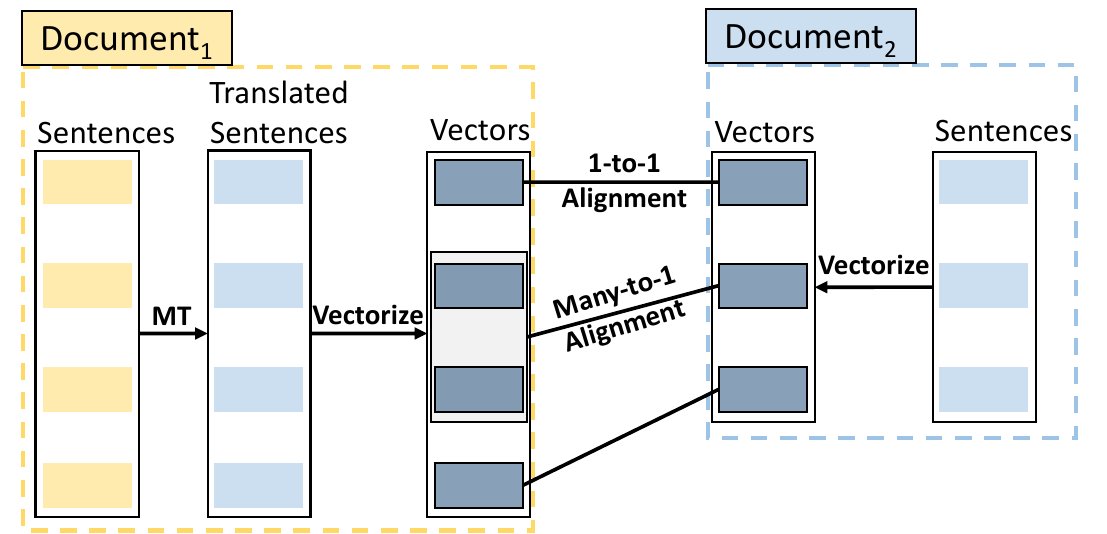}
    \caption{\textbf{The proposed sentence alignment flowchart.} An initial MT system translates one document into the language of the other document. Sentences are converted into embeddings in the same vector space and dynamic programming is applied to extract 1--to--1 and many--to--many alignments.
    }
    \label{fig:sentence_alignment_concept}
\end{figure} 
\label{sec:parallel_sents_extraction}
We extract parallel sentences from parallel documents as illustrated in \Fig{sentence_alignment_concept}.
We first apply an initial MT system to align the languages of the two documents. Secondly, we convert sentences to embeddings in the same vector space. Finally, we apply dynamic programming~\citep{utsuro1994bilingual} to extract only 1--to--1 case but also many--to--many alignments.


\noindent  \textbf{Initial MT System.}
Given a document pair in two different languages, we translate one of them into the other language using an MT system~\citep{sennrich2010mt}. 
To train such a system, we can leverage any existing parallel data in related or even distant domains such as scientific paper domain corpus ASPEC \citep{nakazawa2016aspec} and spoken language domain corpus TED talks, \citep{cettolo2012wit3} or news domain corpus News Commentary \citep{tiedemann2012parallel}.

\noindent \textbf{Similarity Measure.} One key component in the DP algorithm is the scoring function, i.e., similarity measure.
Given two sentences (to be precise, two sequences, each of which can contain more than one sentence) in different languages, the function returns a real number indicating how similar they are.

Existing methods, such as those in \citep{sennrich2010mt}, used sentence-level BLEU scores of machine-translated sentences in the source language against the sentences in the target language as their similarity score, formally:
\begin{equation}
\mathit{Sim}_{\mathit{BLEU}}(s_{i},s^{\prime}_{j}) = \mathit{BLEU}(\mathit{MT}(s_{i}),s^{\prime}_{j}),
\label{eq:bleu}
\end{equation}
where $s_{i}$ and $s^{\prime}_{j}$ are two sentences in different languages.
However, due to the lack of in-domain data, MT systems can give only translations of low quality, and thus the BLEU scores can be misleading, especially for low-resource situations, such as lecture transcripts translation. 

An alternative method is to directly compute the embedding cosine similarity of a given sentence pair \citep{bouamor2018h2}, relying on pre-trained multilingual word embeddings such as word2vec, LASER, or sentBERT to represent sentences in different languages with the same vector space through element-wise addition of word embeddings \citep{mikolov2013distributed}.
However, cross-lingual embeddings are often inaccurate for distant language pairs, especially if they have been pre-trained on data from another domain.

Taking inspiration from both these approaches, we employ MT combined with a cosine similarity of sentence embeddings to measure the similarity of two sentences in different languages, formulated as follows:
\begin{align}
\begin{split}
\mathit{Sim}_{\mathit{EMB}}(s_{i},s^{\prime}_{j})=\cos\left(\mathit{emb}(\mathit{MT}(s_{i})),\mathit{emb}(s^{\prime}_{j})\right)
\label{eq:emb}
\end{split}
\end{align}
We first translate each sentence in the source language document into the target language. We then calculate the sentence embeddings through function $\mathit{emb}(\cdot)$, which averages the embeddings of words in that sentence by pre-trained word vectors,\fn{http://vectors.nlpl.eu/repository} as in the work by \citep{mikolov2013distributed}. 
In practice, we take the average of $Sim_{EMB}(s_i,s^{\prime}_{j})$ and $Sim_{EMB}(s^{\prime}_{j}, s_i)$ to improve the robustness.

\noindent \textbf{DP Algorithm} The pseudocode of DP is shown in \Algo{sentence_alignment}, where $sentences_1$ and $sentences_2$ represent the lists of sentences in the given two respective documents. $Order$ represents the maximum number of sentences we can combine into one. And $similarity(\cdot, \cdot)$ is a function that returns a score showing the similarity of two lines in different languages. The DP algorithm can process not only 1--to--1 alignment, but also many--to--many alignments.

In our scenario, DP is effective because of the monotonic nature of the transcripts, as the corresponding sentences in each pair of documents are roughly in the same order. This is based on the fact that the sentences in lecture transcripts of different languages are in the same order as the professor's speech in the course. Consequently, comparing all pairs of sentences between document pairs is unnecessary~\citep{abdelali2014amara}. 

\begin{algorithm*}[t]
\SetKwInOut{Input}{Input}
\SetKwInOut{Output}{Output}
\Input{$\mathit{sentences_1}$, $\mathit{sentences_2}$, $\mathit{Order}$, $\mathit{similarity()}$}
\Output{$\mathit{path}$}
\ForEach{$\mathit{i}\in\mathit{range(0,len(sentences_1)-1)}$}
{
    \ForEach{$\mathit{j}\in\mathit{range(0,len(sentences_2)-1)}$}
    {
        $\mathit{f[i][j] = -1}$\;
        $\mathit{path[i][j] = (0, 0)}$\;
        \ForEach{$\mathit{\delta_i}\in\mathit{range(0,Order)}$}
        {
            \ForEach{$\mathit{\delta_j}\in\mathit{range(0,Order)}$}
            {
                $\mathit{chunk_{1}=sentences_{1}[i-\delta_i+1]+...+sentences_{1}[i]}$;\\
                $\mathit{chunk_{2}=sentences_{2}[j-\delta_j+1]+...+sentences_{2}[j]}$;\\
                $\mathit{similarity\_score = similarity(chunk_{1}, chunk_{2})}$\;
                $\mathit{score = similarity\_score + f[i-\delta_i][j-\delta_j]}$\;
                \If{$\mathit{score>f[i][j]}$}
                {
                    $\mathit{f[i][j] = score}$\;
                    $\mathit{path[i][j] = (\delta_i,\delta_j)}$\;
                }
            }
        }
    }
}
$\mathit{return\ path}$\;
\caption{Pseudocode for the sentence alignment algorithm. We can recover alignment results from $path$.}
\label{algo:sentence_alignment}
\end{algorithm*}

\subsection{High-quality Evaluation Sets Creation}
\label{sec:evaluation_sets_creation}
A high-quality test set is required to benchmark the performance of educational lecture translation, and a high-quality development set can be useful for tuning MT systems. We create test and development sets by manually filtering the aligned sentence pairs obtained in the previous step.

We first select the document pairs that have a high similarity score under the automatic evaluation similarity criteria. Specifically, we sort all document pairs in descending order of the average similarity of aligned sentence pairs within. 

We then subject these sorted and sentence-aligned pairs to human evaluation to obtain well-aligned test and development sets. As presented in \Algo{testdev}, the target volume of each set ($\mathit{volume}$) and document-level comparability ($\mathit{ratio}$) are the two parameters. In line $2$, while the number of clean sentence pairs is fewer than required, we manually check another document pair. For each document pair, we manually check all sentence pairs and only add the correct ones, as shown in lines $7$ to $10$.  We use the remaining sentence-aligned document pairs for training. Our test, development, and train sets are all constructed at the document level; thus our corpora could be used to evaluate document-level translation in the future, if required.

\begin{algorithm}[t]
\SetKwInOut{Input}{Input}
\SetKwInOut{Output}{Output}
\Input{$\mathit{DocPairs}$, $\mathit{volume}$, $\mathit{ratio}$}
\Output{$\mathit{DocPairs}$, $\mathit{SentencePairs}$}
$\mathit{SentencePairs}\gets\{\}$\;
\While {$|\mathit{SentencePairs}|<\mathit{volume}$}
{
    $\mathit{DocPair}\gets$ pickBestDocPair($\mathit{DocPairs}$)\;
    $\mathit{DocPairs}\gets\mathit{DocPairs}\backslash\{\mathit{DocPair}\}$\;
    $\mathit{CandidatePairs}\gets$ getAlignments($\mathit{DocPair}$)\;
    $\mathit{Correct}\gets$ \{\}\;
    \ForEach{$\mathit{Pair}\in\mathit{CandidatePairs}$}  
    {
        $\mathit{Judge}\gets$ manualEvaluation($\mathit{Pair}$)\;
        \If {$\mathit{Judge}$ == good}
        {
            $\mathit{Correct}\gets\mathit{Correct}\cup\{\mathit{Pair}\}$\;
        }
    }
    \If {$|\mathit{Correct}| > |\mathit{CandidatePairs}| * \mathit{ratio}$}
    {
        $\mathit{SentencePairs}\gets\mathit{SentencePairs}\cup\mathit{Correct}$\;
    }
}
\caption{Document-aware sentence filtering}
\label{algo:testdev}
\end{algorithm}

\section{Corpus Construction Experiments}
\label{sec:corpus_construction_experiments}
We report on the application of our framework to create English--Japanese and English--Chinese Coursera corpus, focusing on settings, performance, and analysis, including 1) cleaning documents, 2) initial MT system creation, 3) sentence alignment results, 4) Coursera corpus creation, and 5) corpus analysis.

\subsection{Document Cleaning Experiments}
\label{sec:cleaning_documents}
For the sentence segmentation settings, we first segment paragraphs with a full-stop (``.''), exclamation (``!''), and question marks (``?'') in Latin encoding and their full-width counterparts in UTF-8 followed by a space or the end of the line. We then tokenize sentences, using Juman++ \citep{jumanpp1}\fn{\url{https://github.com/ku-nlp/jumanpp}} for Japanese, NLTK tool\fn{\url{https://www.nltk.org}} for English and StanfordCoreNLP\fn{\url{https://stanfordnlp.github.io/CoreNLP/}} for Chinese.

For the language detection settings, we calculate the number of English and Japanese characters to judge whether the given document is in Japanese or English. More specifically, we define a set of characters, \texttt{EnglishChar}, with ``a'' to ``z'' and ``A'' to ``Z,'' and another set of characters, \texttt{JapaneseChar}, with \texttt{Hiragana} and \texttt{Katakana} characters. Compared with the langdetect tool~\citep{raffel2019exploring}, \hl{on an evaluation set of 100 manually annotated samples, the rule-based method classified the language of all documents correctly and the langdetect tool correctly classified 99 of them.} For English--Chinese, we rely on the langdetect tool to filter out data in other languages.

\subsection{Initial MT System Creation}
\label{sec:initial_mt_system_creation}

We use TED Talks corpus \citep{cettolo2012wit3} and ASPEC corpus \citep{nakazawa2016aspec}\fn{We selected the best 1.0 million sentence pairs according to the pre-computed score in the provided corpus for each sentence pair.} to build the English--Japanese MT system, use TED talks and News Commentary~\cite{tiedemann2012parallel} to build the English--Chinese MT system.  \Tab{aspected} gives the statistics of these three corpora.

To train the MT models, we leverage the smaller TED Talks corpus and the larger ASPEC or News Commentary corpus through fine-tuning and mixed fine-tuning approaches \cite{raj17}. When performing mixed fine-tuning on the concatenation of both corpora, the TED Ja--En and TED Zh--En corpora were oversampled to match the size of the ASPEC Ja--En and News Commentary Zh--En corpora, respectively.
NMT models are trained using tensor2tensor framework with its default hyperparameters as shown in \Sec{mtsettings}.

We evaluate these MT models through a test set from the TED dataset through the BLEU score \cite{papineni}, since we do not have a test set for the target domain yet, i.e., Coursera. \Tab{initmt} gives the results where ``A,'' ``T,'' and ``N,'' stand for ASPEC, TED, and News Commentary, ``AT'' or ``NT'' stands for the balanced mixture of two datasets. ``{\ra}'' means fine-tuning on the right-hand side data.

We found that the mixed fine-tuning strategy yields the best results for both Japanese\ra English and Chinese\ra English translations. We, therefore, use these models during the sentence alignment procedure. 

\begin{table}[t]
    \centering
    \caption{Number of sentence pairs in each English--Japanese and English--Chinese corpus to train initial MT systems.}
    \resizebox{\linewidth}{!}{
    \begin{tabular}{crrrr}
    \toprule
    \multirow{2}{*}{\diagbox{Split}{Dataset}} & \multicolumn{2}{c}{English--Japanese} & \multicolumn{2}{c}{English--Chinese} \\
    & ASPEC & TED & News & TED \\ 
    \toprule
    Train & 1.0M & 223k & 321k &215k \\ 
    Development& 1,790 & 1,354 & - & 1,261\\ 
    Test & 1,812 & 1,194 & - & 1,064 \\ \bottomrule
    \end{tabular}
    }
    \label{tab:aspected}
\end{table}

\begin{table}[thb]
    \centering
    \caption{BLEU scores of \task{Ja}{En} and \task{Zh}{En} translation directions on TED test set. ``A'' represents the ASPEC train set, ``N'' represents the News Commentary train set, and ``T'' represents the TED train set. ``\ra'' indicates fine-tuning. The best result is marked \textbf{bold}.}
    \small
    \begin{tabular}{cr}
        \toprule
        \multicolumn{2}{c}{\textbf{Japanese\ra English}} \\
        Training path& BLEU score\\ \midrule
        A & 4.1  \\ 
        T & 12.2  \\
        AT & 14.6 \\
        A {\ra} T& 13.9 \\ 
        A {\ra} AT & \textbf{15.0} \\
        \bottomrule
        \multicolumn{2}{c}{\textbf{Chinese\ra English}} \\
        Training path & BLEU\\
        \midrule
        N & 11.1  \\ 
        N {\ra} NT & \textbf{21.5} \\ \bottomrule
    \end{tabular}
    \label{tab:initmt}
\end{table}

\subsection{\hl{Sentence Alignment Experiments}}
\label{sec:sentence_alignment_experiments}
\hl{We compare our proposed sentence alignment methods with the BLEU-score based method~\citep{sennrich2010mt}, BERTScore~\citep{bert-score} based method, LASER universal embedding based methods~\citep{chaudhary2019low, thompson-koehn-2019-vecalign}, and sentBERT embedding~\citep{reimers-2020-multilingual-sentence-bert} based method. We measure the alignment $F_1$ score on four manually annotated document pairs (2 English--Japanese and 2 English--Chinese) with a total of 426 lines.}

\subsubsection{\hl{Sentence Alignment Methods \mycomment{This section describes baseline methods and proposed methods.}}}
\label{sec:sentence_alignment_methods}
\hl{We try the combinations of these variables in the alignment methods: 1) embedding methods including word2vec, LASER, sentBERT, BERT, or w/o embedding for BLEU-score based method, 2) similarity measurement: cosine similarity, L2 distance, BERTScore, or BLEU-score, and 3) alignment algorithms including greedy, minimum-cost flow, and DP.}

\hl{Regarding the embedding methods, word2vec models we used are from NLPL word embedding repository}.\fn{\url{http://vectors.nlpl.eu/repository}, ID 35, 40, 53 for Chinese, English, and Japanese, respectively} \hl{We convert words in one sentence into embeddings and average the word embeddings as sentence embeddings. We used RoBERTa-large\fn{\url{https://github.com/Tiiiger/bert_score}} for BERTScore. The LASER model we used is a multilingual BiLSTM model}\fn{\url{dl.fbaipublicfiles.com/laser/models/bilstm.93langs.2018-12-26.pt}} \hl{and the sentBERT model we used is a multilingual MPNet model.}\fn{\url{https://huggingface.co/sentence-transformers/all-mpnet-base-v2}} \hl{The multilingual sentence-to-vector models can convert sentences in different languages into embeddings. Note that the vector spaces are aligned across languages for LASER but not aligned for sentBERT.}

\hl{
Concerning the similarity measurement, 
for the BLEU score which is the higher the better, we use the NLTK toolkit to calculate sentence BLEU score with the default setting. 
BERTScore, the higher the better, leverages the pre-trained contextual embeddings from a BERT-like model and matches words in candidate and reference sentences by cosine similarity as defined in~\citep{bert-score} and we use the $F_1$ score of BERTScore. We translate sentences in Japanese and Chinese into English before applying BERTScore.}

In the case of cosine similarity, the higher the better, is defined as:

\begin{equation}
\cos ({\bf v},{\bf v^{\prime}})= \frac{ \sum_{i=1}^{D}{{\bf v}_i{\bf v^{\prime}}_i} }{ \sqrt{\sum_{i=1}^{D}{({\bf v}_i)^2}} \sqrt{\sum_{i=1}^{D}{({\bf v^{\prime}}_i)^2}} }
\label{eq:cosine_similarity}
\end{equation}
where ${\bf v}$ and ${\bf v^{\prime}}$ are two sentence embeddings from $Document_1$ and $Document_2$ with dimension $D$ and ${\bf v}_i$ or ${\bf v^{\prime}}_i$ is the $i_{th}$ dimension of embedding ${\bf v}$ or ${\bf v^{\prime}}$.

\hl{Finally, L2 distance, the lower the better, we define it as below. To ensure consistency in the analogy with other measurements, in which higher values are better, we take the negative of L2 distance in practice.}
\hl{
\begin{equation}
 Distance_{L2}\left( {\bf v},{\bf v}^{\prime}\right)   = \sqrt {\sum_{i=1}^{D}  \left( {\bf v}_{i}-{\bf v}^{\prime}_{i}\right)^2 } 
\label{eq:l2distance}
\end{equation}
}

\hl{Besides the DP algorithms as shown in \Algo{sentence_alignment}}\hl{, we also try two other algorithms: greedy and minimum-cost flow.
The greedy algorithm is simple and implemented as follows: for each sentence in the source language $s_i$, we find a sentence in the target language $s^{\prime}_j$ that maximizes $Similarity(s_i,s^{\prime}_j)$.
The set $S$ of matched pairs $(i,j)$ for all the extracted sentences in the source language $s_i$ with a corresponding sentence in the target language $s^{\prime}_j$ can be represented as follows:}
\hl{
\begin{equation}
S=\{(i,\argmax_{j}[Similarity(s_i, s^{\prime}_j)]\}_{i=1}^{N}
\label{eq:greedy_method}
\end{equation}
}
\hl{Note that using the greedy algorithm, it is possible that for two pairs $(s_i, s^{\prime}_j)$ and $(s_k, s^{\prime}_l)$ with $s_i\neq s_k$ and $s^{\prime}_j=s^{\prime}_l$. This method is suitable for mining parallel sentences from large-scale web corpora, e.g. BUCC bitext mining task using LASER.\fn{\url{https://github.com/facebookresearch/LASER/blob/main/tasks/bucc}} 
}

\hl{An improved version of the greedy algorithm is using the minimum cost maximum flow algorithm\footnote{\url{https://en.wikipedia.org/wiki/Minimum-cost_flow_problem}} to achieve 1--to--1 global best match, where the algorithm gives maximum total similarity scores allowing only 1--to--1 match. For every two pairs  $(s_i, s^{\prime}_j)$ and $(s_k, s^{\prime}_l)$ in $S$, $s_i\neq s_k$, $s^{\prime}_j\neq s^{\prime}_l$ always holds. The graph is constructed as follows; an example is shown in \Fig{mincostmaxflow}:
\begin{itemize}
    \item from the source point $A$ to the all sentences in the source language $s_i$, add edges with $(capability,cost)=(1,0)$
    \item from all sentences in the target language $s^{\prime}_j$ to the target point $B$, add edges with $(capability,cost)=(1,0)$
    \item from all sentences in the source language $s_i$ to all sentences in the target language $s^{\prime}_j$, add edges with $(capability,cost)=(1,-similarity(s_i, s^{\prime}_j))$
\end{itemize}
We then send an infinite amount of flow from source vertex $A$ to sink vertex $B$ and obtain the residual network. We select sentence pair $(s_i,s^{\prime}_j)$ if the remaining capability of edge $(s_i,s^{\prime}_j)$ is zero. By the property of this algorithm, it is more suitable for extracting parallel sentences from two comparable documents instead of corpora mining from the web.
\begin{figure}[t]
    \centering
    \includegraphics[width=0.7\linewidth]{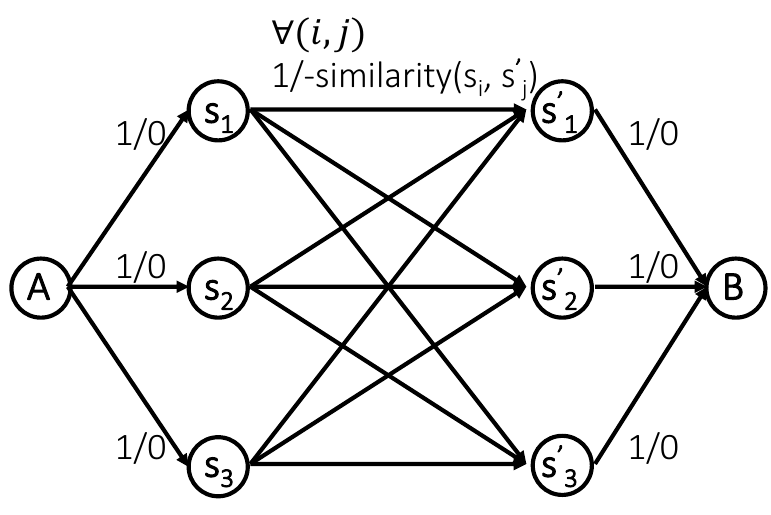}
    \caption{\textbf{The graph for the minimum-cost maxflow algorithm.} A represents the source and B represents the sink, $s_i$ are the points for sentences in the source language and $s^{\prime}_j$ are the points for sentences in the target language. Each edge is with $capability/cost$ information.}
    \label{fig:mincostmaxflow}
\end{figure} }


\begin{table}[thb]
    \centering
    \caption{\hl{The statistical information of $2$ English--Japanese and $2$ English--Chinese annotated document pairs for sentence alignment evaluation.}}
    \small
    \begin{tabular}{crrr}
        \toprule
        \hl{\#Doc} & \hl{\#En lines} & \hl{\#Ja lines} & \hl{\#Zh lines} \\
        \midrule
        \hl{1} & \hl{75} & \hl{91} & \hl{-} \\ 
        \hl{2} & \hl{52} & \hl{63} & \hl{-} \\ 
        \hl{3} & \hl{34} & \hl{-} & \hl{22} \\ 
        \hl{4} & \hl{47} & \hl{-} & \hl{42} \\ 
        \bottomrule
        \hl{Total} & \hl{208} & \hl{154} & \hl{64} \\ 
        \bottomrule
    \end{tabular}
    \label{tab:document_pairs_for_alignment}
\end{table}
\begin{table*}[t]
    \centering
    \caption{\hl{\textbf{Average $F_1$ scores of different sentence alignment methods.} Tested on four manually annotated document pairs. The best result is marked \textbf{bold}. We also show $F_1$ scores of two English--Japanese document pairs $F_1^{En-Ja}$ and $F_1$ scores of two English--Chinese document pairs $F_1^{En-Zh}$ to the footer of each $F_1$ score.}}
    \begin{tabular}{ccc|lll}
        \toprule
        \multirow{2}{*}{\hl{Model}} &  \multirow{2}{*}{\hl{With MT}}  & \multirow{2}{*}{\hl{Similarity}} &\multicolumn{3}{c}{\hl{Algorithm}} \\
        & & & \hl{Greedy} & \hl{Min-cost flow} & \hl{DP} \\
        \midrule
        \hl{\underline{\textit{Baseline:}}} & & & & \\
        \hl{-} & \hl{Yes} & \hl{BLEU} & \hl{$0.74_{0.76/0.72}$} & \hl{$0.82_{0.79/0.85}$} & \hl{$0.93_{0.98/0.89}$} \\
        \hl{RoBERTa} & \hl{Yes} & \hl{BERTScore} & \hl{$0.70_{0.74/0.66}$} & \hl{$0.81_{0.79/0.84}$} & \hl{$0.89_{0.97/0.82}$} \\
        \hl{LASER} & \hl{No} & \hl{Cosine} & \hl{$0.75_{0.77/0.73}$} & \hl{$0.82_{0.78/0.85}$} & \hl{$0.90_{0.98/0.82}$} \\
        \hl{LASER} & \hl{No} & \hl{Distance} & \hl{$0.75_{0.77/0.73}$} & \hl{$0.82_{0.78/0.85}$} & \hl{$0.83_{0.91/0.76}$} \\
        \hl{sentBERT} & \hl{No} &\hl{Cosine} & \hl{$0.12_{0.08/0.16}$} & \hl{$0.18_{0.10/0.25}$} & \hl{$0.52_{0.41/0.64}$} \\
        \hl{sentBERT} & \hl{No} & \hl{Distance} & \hl{$0.12_{0.08/0.16}$} & \hl{$0.18_{0.10/0.25}$} & \hl{$0.25_{0.18/0.32}$} \\
        \midrule
        \hl{\underline{\textit{Proposed:}}} & & & & \\
        \hl{word2vec} & \hl{Yes} &\hl{Cosine} & \hl{$0.65_{0.73/0.57}$} & \hl{$0.79_{0.76/0.82}$} & \hl{$\boldsymbol{0.96}_{0.97/0.95}$} \\
        \bottomrule
    \end{tabular}
    \label{tab:alignment_performance}
\end{table*}
\hl{\subsubsection{Sentence Alignment Benchmark}
\label{sec:sentence_alignment_benchmark}
We manually annotated gold alignments of four document pairs and the statistical information is shown in \Tab{document_pairs_for_alignment}. There are 2 English--Japanese and 2 English--Chinese document pairs containing a total of 426 sentences of 208 English sentences, 154 Japanese sentences, and 64 Chinese sentences. In detail, they contain 162 ($84\%$) 1--to--1 alignment cases, 8 ($4\%$) many--to--1 cases, 21 ($11\%$) 1--to--many cases, and 1 ($1\%$) many--to--many alignment cases, which requires the ability to combine multiple lines into one.
}

\subsubsection{Sentence Alignment Performance}
\label{sec:sentence_alignment_benchmark}
We measure the performance of alignment methods through macro $F_1$ score on four document pairs. For each document pair, the number of manually annotated alignment pairs is defined as $Total\_Pairs$, the number of extracted alignment pairs is defined as $Extracted\_Pairs$, and the number of correctly extracted pairs is defined as $Correct\_Pairs$. We define $Precision$ as $\frac{Correct\_Pairs}{Extracted\_Pairs}$, $Recall$ as $\frac{Correct\_Pairs}{Total\_Pairs}$, and $F_1$ score as $\frac{2*Precision*Recall}{Precision+Recall}$.

\hl{The results of the combinations of embeddings, algorithms, and similarity measurements are shown in \Tab{alignment_performance}. Our proposed method using word2vec as embedding method, cosine similarity as similarity measurement, and DP algorithm achieved the best macro average $F_1$ score of 96\% on four document pairs, which is higher than the performance of the BLEU-based method~\citep{sennrich2010mt}, BERTScore-based method~\citep{bert-score}, original LASER method with greedy algorithm~\citep{chaudhary2019low} for BUCC task that is only 75\% and even higher than the LASER with DP~\citep{thompson-koehn-2019-vecalign} method that is 90\%.
Additionally, we can observe the following:
\begin{itemize}
    \item In our situation where there are no crossing alignments, the DP algorithm achieved much better results than greedy, or minimum-cost flow, because it can process the 1--to--many, many--to--1 and many--to--many sentence alignment cases.
    \item MT-based methods, including word2vec, BLEU-based and BERTScore, are better than multilingual embedding based methods, including the LASER and sentBERT based methods. In the English--Japanese and English--Chinese scenarios where relatively good initial MT systems exist, it is better to first translate the sentences in the source language into sentences in the target language and measure the similarity using the two sentences in the same language. The performance of the sentBERT model is low because the vector spaces are not aligned across languages.
    \item Using cosine similarity gives slightly better performance than the pairwise distance that is used in the LASER method.
\end{itemize} }

\begin{table}[thb]
    \centering
    \caption{The size of English--Japanese Coursera parallel dataset.}
    \resizebox{\linewidth}{!}{
    \begin{tabular}{crrr}
        \toprule
        & \makecell{\# of \\ document pairs} &  \makecell{\# of \\ aligned lines} & \makecell{\# of \\deleted lines} \\
        \toprule
        Test & 50  & 2,068 & 140  \\ 
        Development& 16  & 555 &  88\\
        Train & 818 & 50,543 & - \\ 
        \bottomrule
    \end{tabular}
    }
    \label{tab:testdevset}
\medskip
    \centering
    \caption{The size of English--Chinese Coursera parallel dataset.}
    \resizebox{\linewidth}{!}{
    \begin{tabular}{crrr}
        \toprule
        & \makecell{\# of \\ document pairs} &  \makecell{\# of \\ aligned lines} & \makecell{\# of \\deleted lines} \\
        \toprule
        Test & 90  & 2,009 & 83\\ 
        Development& 34  & 865 &  43\\ 
        Train & 997 & 40,074 & - \\ 
        \bottomrule
    \end{tabular}
    }
    \label{tab:testdevset_zhen}
\end{table}

\subsection{Coursera Dataset Creation} 
\label{sec:coursera_dataset_creation}
We introduce the created corpora and detailed settings in this section. With the proposed alignment algorithm, we obtained a total of 53,394 pairs of sentences from 884 document pairs for English--Japanese and a total of 43,074 pairs of sentences from 1,121 document pairs for English--Chinese. 

Test and development sets are created through manually filtering\fn{The checker is a native Chinese speaker, has the N1 certification (the highest level) of the Japanese Language Proficiency Test, and got 99 points in TOEFL iBT.} as introduced in \Sec{evaluation_sets_creation}. We set 2,000 and 500 sentences for English--Japanese and 2,000 and 850 for English--Chinese as the target $\mathit{volume}$ for the test and development sets, respectively, and set $\mathit{ratio}=0.50$.
As shown in \Tab{testdevset}, for English--Japanese, a total of 2,851 sentence pairs drawn from 66 documents were manually judged and approximately 8.0\% of them ((140+88)/2,851) were filtered out. We did not perform manual filtering for the train set due to its large scale.

As a result, the train, development, and test sets in the English--Japanese corpus contain \textbf{50,543}, \textbf{555}, and \textbf{2,086} parallel sentences, respectively.
For English--Chinese, with the same method, the train, development, and test sets contain \textbf{40,074}, \textbf{865}, and \textbf{2,009} parallel sentences, respectively.

\subsection{Corpus Analysis} 
\label{sec:corpus_analysis}
We show the sentence length analysis and similarity with other corpora of our created Coursera corpora.

The sentence length information is presented in \Tab{length-distribution} and \Tab{length-distribution_zhen}, with the tokenization methods introduced in \Sec{cleaning_documents}. 
For the English--Japanese Coursera corpus, the average sentence length is between ASPEC and TED and relatively closer to TED. For the English--Chinese Coursera corpus, the average sentence length is between that of News Commentary and TED corpora and is closer to News Commentary.

We measure the similarity with other corpora using the language model (LM). Specifically, we trained 4-gram LMs on the lower-cased version of the English side of each train set. We then computed the per-token log-likelihood of train sets with each of these LMs. 
As shown in \Tab{datasetsimilarity}, the three English--Japanese datasets are visibly distant from each other. Nevertheless, TED seems relatively more exploitable than ASPEC for helping to translate Coursera datasets, presumably because they comprise a spoken language, unlike ASPEC.
As given by \Tab{datasetsimilarity_zhen}, when tested on the LM trained on the Coursera corpus, the TED corpus achieves a higher likelihood than that of the News commentary corpus, suggesting that the TED corpus is more similar to the Coursera corpus than the News commentary corpus.

\begin{table}[thb]
    \centering
    \caption{Sentence length of English--Japanese train set.}
    \resizebox{\linewidth}{!}{
    \begin{tabular}{crr}
        \toprule
        \multirow{2}{*}{Dataset} &\multicolumn{1}{c}{English} &\multicolumn{1}{c}{Japanese} \\
        &Mean / Median / s.d. &Mean / Median / s.d. \\ \toprule
        ASPEC		& 25.4 / 23 / 11.4 & 27.5 / 20 / 12.0 \\ 
        TED		& 20.4 / 17 / 13.9 & 19.8 / 16 / 14.1 \\ 
        Coursera	& 21.1 / 19 / 11.1 & 22.2 / 20 / 11.8 \\ \bottomrule
    \end{tabular}
    }
    \label{tab:length-distribution}
\medskip
    \centering
    \caption{Sentence length of English--Chinese train set.}
    \resizebox{\linewidth}{!}{
    \begin{tabular}{crr}
        \toprule
        \multirow{2}{*}{Dataset} &\multicolumn{1}{c}{English} &\multicolumn{1}{c}{Chinese} \\
        &Mean / Median / s.d. &Mean / Median / s.d. \\ \toprule
        News commentary & 26.0 / 25 / 12.7 & 24.7 / 23 / 12.2 \\ 
        TED		& 20.5 / 17 / 15.3 & 18.9 / 15 / 14.5 \\ 
        Coursera	& 25.7 / 22 / 25.7 & 24.2 / 20 / 22.2 \\ \bottomrule
    \end{tabular}
    }
    \label{tab:length-distribution_zhen}
\end{table}

\begin{table}[thb]
    \centering
    \caption{Average per-token log-likelihood of En--Ja datasets.}
    \resizebox{\linewidth}{!}{
    \begin{tabular}{crrr}
        \toprule
        \diagbox{LM}{Corpus} & ASPEC & TED & Coursera\\ \toprule
        ASPEC		& -1.157  & -3.543  & -3.364  \\
        TED		& -3.468  & -1.158  & -2.570  \\ 
        Coursera	& -3.482  & -2.800  & -0.790  \\ \bottomrule
    \end{tabular}
    }
    \label{tab:datasetsimilarity}
\medskip
    \centering
    \caption{Average per-token log-likelihood of En--Zh datasets.}
    \resizebox{\linewidth}{!}{
    \begin{tabular}{crrr}
        \toprule
        \diagbox{LM}{Corpus} & News commentary& TED & Coursera\\ \toprule
        News commentary& -1.001  & -2.516 & -2.635 \\ 
        TED		&  -2.775 & -1.055 &  -2.375 \\ 
        Coursera	&  -2.976 &  -2.542 &  -0.824 \\ \bottomrule
        \end{tabular}
        }
        \label{tab:datasetsimilarity_zhen}
\end{table}

\section{MT Experiments}
\label{sec:mt_experiments}
\hl{
In this section, we show the effectiveness of the in-domain parallel corpora created by the proposed method through MT experiments. Specifically, we apply the created lecture domain parallel corpora to creating an MT system, combining it with larger-scale out-of-domain corpora. To leverage small in-domain and large out-of-domain datasets efficiently, we modify the two-stage fine-tuning method to a multistage fine-tuning method.
}

\subsection{Datasets}
\label{sec:datasets}
We leveraged both the in-domain Coursera corpora and out-of-domain corpora in the English--Japanese and English--Chinese MT experiments.
The information of the corpora is shown in Tables~\ref{tab:aspected},~\ref{tab:testdevset}, and~\ref{tab:testdevset_zhen}.

We used these abbreviations for the datasets: in the English--Japanese MT experiments, ``A'' stands for ASPEC train set of 1.0 million lines, ``T'' stands for TED Talks train set of 0.2 million lines, and ``C'' for Coursera train set of 50 thousand lines. In the English--Chinese MT experiments, we refer to the News Commentary train set with 0.3 million lines as ``N,'' TED train set with 0.2 million lines as ``T,'' and the Coursera train set with 40 thousand lines as ``C.''

When performing dataset combinations, we always oversample the smaller ones to match the size of the largest ones. We denote the concatenated corpus by a concatenation of the letters representing them: e.g., ``AT'' for the mixture of ASPEC data with five times oversampled TED data, and ``ATC'' for the concatenation of ASPEC with five times oversampled TED data and 25 times oversampled Coursera data. We use \ra$D$ to represent fine-tuning the model on the dataset $D$, e.g. ``A''\ra ``AT'' means that we first train the model on ASPEC and then fine-tune it on the mixture of ASPEC and TED with oversampling.

All the datasets are tokenized into subwords using Byte-Pair-Encoding (BPE) \citep{sennrich-etal-2016-neural} method. For English--Japanese experiments, we created a shared Japanese and English subword vocabulary from the ASPEC and TED train sets with oversampling, with 32,000 merge operations. For English--Chinese experiments, we created a shared Chinese and English subword vocabulary from the News Commentary and TED train sets with oversampling, with 32,000 merge operations. These vocabularies are used for all experiments.

\subsection{Model and Training Settings}
\label{sec:mtsettings}
We used the tensor2tensor framework \citep{tensor2tensor}\fn{\url{https://github.com/tensorflow/tensor2tensor}, version \href{https://github.com/tensorflow/tensor2tensor/releases/tag/v1.14.0}{1.14.0}.} with ``\emph{transformer\_big}'' architecture and default settings for this architecture, such as $dropout=0.2$, $attention\ dropout=0.1$, $optimizer=adam$ with $beta1=0.9$, $beta2=0.997$.

During training, we used eight Tesla V100 32GB GPUs with a batch size of 4,096 subword tokens in all the experiments. We used early stopping on token-level BLEU score computed on the development set: the training process stops if the score shows no gain larger than 0.1 for 10,000 steps. When fine-tuning the model on a different dataset, we always resumed the training process from the last checkpoint in the previous stage.

During inference, we used the average of the last ten checkpoints and decoded the test sets with a beam size of 4 and a length penalty $\alpha$ of 0.6 consistently across all the models. 
We calculate the BLEU scores for the decoded text and target text using the SacreBLEU tool.\fn{\url{https://github.com/mjpost/sacreBLEU}} 

\begin{table*}[t]
    \centering
    \caption{\hl{\textbf{BLEU of English--Japanese translation for all domain adaptation methods on Coursera test set.}
    The training path shows the datasets used in each fine-tuning stage. The best score in each direction is marked \textbf{bold}.}}
    \begin{tabular}{ccrr}
        \toprule
        \hl{Method} & \hl{Training path} & \hl{Japanese}\ra \hl{English} & \hl{English}\ra \hl{Japanese}\\  \toprule
        \underline{\textit{Baseline:}}\\
        \hl{Out-of-domain data} & \hl{A} & \hl{14.1} & \hl{10.4} \\
        \hl{Out-of-domain data} & \hl{T} & \hl{16.7} & \hl{8.6} \\
        \hl{In-domain data only} & \hl{C} & \hl{6.8} & \hl{7.7} \\
        \hl{Mixed out-of-domain data} & \hl{AT} & \hl{24.9} & \hl{13.2} \\
        \hl{Mixed data} & \hl{ATC} & \hl{25.6} & \hl{17.3} \\
        \hl{Two-stage fine-tuning} & \hl{ATC}\ra \hl{C} & \hl{23.8} & \hl{18.1} \\
        \hl{Mixed fine-tuning} & \hl{A}\ra \hl{ATC}& \hl{27.8} & \hl{17.9} \\
        \midrule
        \underline{\textit{Proposed:}}\\
        \hl{\textbf{Multistage fine-tuning}} & \hl{A}\ra \hl{AT}\ra \hl{ATC} & \hl{\textbf{27.9}} & \hl{\textbf{18.5}} \\
        \bottomrule
    \end{tabular}
    \label{tab:selected_results}
\end{table*}

 \begin{table*}[thb]
    \centering
    \caption{\hl{\textbf{BLEU of English--Chinese translation for all domain adaptation methods on Coursera test set.}
    The training path shows the datasets used in each fine-tuning stage. The best score in each direction is marked \textbf{bold}.}}
    \begin{tabular}{ccrr}
        \toprule
        \hl{Method} & \hl{Training path} & \hl{Chinese}\ra \hl{English} & \hl{English}\ra \hl{Chinese}\\  \toprule
        \underline{\textit{Baseline:}}\\
        \hl{Out-of-domain data} & \hl{N} & \hl{18.3} & \hl{13.3} \\
        \hl{Out-of-domain data} & \hl{T} & \hl{15.5} & \hl{11.6} \\
        \hl{In-domain data only} & \hl{C} & \hl{14.8} & \hl{14.5} \\
        \hl{Mixed out-of-domain data} & \hl{NT} & \hl{21.6} & \hl{18.4} \\
        \hl{Mixed data} & \hl{NTC} & \hl{28.2} & \hl{26.6} \\
        \hl{Two-stage fine-tuning} & \hl{NTC}\ra \hl{C} & \hl{27.0} & \hl{\textbf{27.1}} \\
        \hl{Mixed fine-tuning} & \hl{N}\ra \hl{NTC}& \hl{\textbf{29.5}} & \hl{26.8} \\
        \midrule
        \underline{\textit{Proposed:}}\\
        \hl{\textbf{Multistage fine-tuning}} & \hl{N}\ra \hl{NT}\ra \hl{NTC} & \hl{\textbf{29.5}} & \hl{\textbf{27.1}} \\
        \bottomrule
    \end{tabular}
    \label{tab:selected_results_zh_en}
\end{table*}

\begin{table*}[tbh]
    \centering
    \caption{BLEU scores for all the multistage training options examined in our Japanese--English MT experiment.  Models A1--A16 and B2--B16 represent all 31 ($=2^{5}-1$) sub-paths of the A{\ra}AT{\ra}ATC{\ra}TC{\ra}C flow. Bold indicates the \textbf{initial training}, and red, blue, and gray cells indicate \colorbox{cR}{inflation}, \colorbox{cB}{deflation}, and \colorbox{cK}{replacement} of the training data, respectively.}

\scalebox{0.85}[0.85]
{
\begin{tabular}{|c|ccccc|r|r|p{0mm}|c|cccc|r|r|}
\cline{1-8}\cline{10-16}
  ID &\multicolumn{5}{c|}{Training schedule} &\task{Ja}{En} &\task{En}{Ja}
&&ID &\multicolumn{4}{c|}{Training schedule} &\task{Ja}{En} &\task{En}{Ja}\\
\cline{1-8}\cline{10-16}

A1	&\bf A	&	&	&	&	&14.1 	&10.4 	&	&	&	&	&	&	&	&\\
A2	&\bf A	&\bgR{AT}	&	&	&	&26.1 	&14.0 	&	&B2	&\bf AT	&	&	&	&24.9 	&13.2\\
A3	&\bf A	&\bgR{AT}	&\bgR{AT}C	&	&	&\bf 27.9 	&18.5 	&	&B3	&\bf AT	&\bgR{ATC}	&	&	&27.4 	&18.1\\
A4	&\bf A	&\bgR{AT}	&\bgR{AT}C	&\bgB{TC}	&	&27.2 	&17.5 	&	&B4	&\bf AT	&\bgR{ATC}	&\bgB{TC}	&	&26.3 	&17.3\\
A5	&\bf A	&\bgR{AT}	&\bgR{AT}C	&\bgB{TC}	&\bgB{C}	&25.9 	&18.6 	&	&B5	&\bf AT	&\bgR{ATC}	&\bgB{TC}	&\bgB{C}	&25.0 	&18.1\\
A6	&\bf A	&\bgR{AT}	&\bgR{AT}C	&	&\bgB{C}	&25.8 	&\bf 18.8 	&	&B6	&\bf AT	&\bgR{ATC}	&	&\bgB{C}	&25.4 	&18.3\\
A7	&\bf A	&\bgR{AT}	&	&\bgK{TC}	&	&27.8 	&18.4 	&	&B7	&\bf AT	&	&\bgK{TC}	&	&26.8 	&17.9\\
A8	&\bf A	&\bgR{AT}	&	&\bgK{TC}	&\bgB{C}	&25.7 	&18.3 	&	&B8	&\bf AT	&	&\bgK{TC}	&\bgB{C}	&24.9 	&17.7\\
A9	&\bf A	&\bgR{AT}	&	&	&\bgK{C}	&25.3 	&18.2 	&	&B9	&\bf AT	&	&	&\bgK{C}	&24.8 	&18.3\\
A10	&\bf A	&	&\bgR{ATC}	&	&	&27.8 	&17.9 	&	&B10	&	&\bf ATC	&	&	&25.6 	&17.3\\
A11	&\bf A	&	&\bgR{ATC}	&\bgB{TC}	&	&27.1 	&17.4 	&	&B11	&	&\bf ATC	&\bgB{TC}	&	&24.4 	&16.8\\
A12	&\bf A	&	&\bgR{ATC}	&\bgB{TC}	&\bgB{C}	&25.5 	&18.4 	&	&B12	&	&\bf ATC	&\bgB{TC}	&\bgB{C}	&23.4 	&17.6\\
A13	&\bf A	&	&\bgR{ATC}	&	&\bgB{C}	&26.3 	&18.2 	&	&B13	&	&\bf ATC	&	&\bgB{C}	&23.8 	&18.1\\
A14	&\bf A	&	&	&\bgK{TC}	&	&27.3 	&18.0 	&	&B14	&	&	&\bf TC	&	&18.7 	&13.3\\
A15	&\bf A	&	&	&\bgK{TC}	&\bgB{C}	&25.1 	&18.1 	&	&B15	&	&	&\bf TC	&\bgB{C}	&18.8 	&14.0\\
A16	&\bf A	&	&	&	&\bgK{C}	&23.2 	&17.5 	&	&B16	&	&	&	&\bf C	&6.8 	&7.7\\\cline{1-8}\cline{10-16}
C3	&\bf A	&\bgR{AT}	&\bgB{T}	&	&	&24.1 	&12.1 	&	&D3	&\bf AT	&\bgB{T}	&	&	&22.6 	&11.5\\
C4	&\bf A	&\bgR{AT}	&\bgB{T}	&\bgR{TC}	&	&25.9 	&17.4 	&	&D4	&\bf AT	&\bgB{T}	&\bgR{TC}	&	&25.5 	&16.5\\
C5	&\bf A	&\bgR{AT}	&\bgB{T}	&\bgR{TC}	&\bgB{C}	&25.3 	&18.1 	&	&D5	&\bf AT	&\bgB{T}	&\bgR{TC}	&\bgB{C}	&24.4 	&17.3\\
C6	&\bf A	&\bgR{AT}	&\bgB{T}	&	&\bgK{C}	&24.6 	&17.5 	&	&D6	&\bf AT	&\bgB{T}	&	&\bgK{C}	&24.0 	&18.1\\
C10	&\bf A	&	&\bgK{T}	&	&	&24.0 	&11.9 	&	&D10	&	&\bf T	&	&	&16.7 	&8.6\\
C11	&\bf A	&	&\bgK{T}	&\bgR{TC}	&	&26.2 	&17.1 	&	&D11	&	&\bf T	&\bgR{TC}	&	&21.4 	&14.1\\
C12	&\bf A	&	&\bgK{T}	&\bgR{TC}	&\bgB{C}	&24.9 	&17.7 	&	&D12	&	&\bf T	&\bgR{TC}	&\bgB{C}	&21.1 	&15.9\\
C13	&\bf A	&	&\bgK{T}	&	&\bgK{C}	&24.3 	&17.7 	&	&D13	&	&\bf T	&	&\bgK{C}	&20.7 	&16.0\\\cline{1-8}\cline{10-16}
E14	&\bf A	&	&	&\bgR{AC}	&	&24.8 	&18.1 	&	&F14	&	&	&\bf AC	&	&19.5 	&15.2\\
E15	&\bf A	&	&	&\bgR{AC}	&\bgB{C}	&23.7 	&17.9 	&	&F15	&	&	&\bf AC	&\bgB{C}	&19.4 	&15.4\\\cline{1-8}\cline{10-16}

\end{tabular}
}

    \label{tab:results}
\end{table*}

\hl{
\subsection{Domain Adaptation Methods}
\label{sec:domain_adaptation_methods}
We describe domain adaptation methods used in the experiments when combining the smaller in-domain Coursera corpus with larger out-of-domain corpora.
}

\hl{
The simplest baseline is to mix multiple datasets (with oversampling) without fine-tuning. However, the performance of NMT is extremely sensitive to the domain of the dataset and to the order in which datasets are included in the training stages. As such, instead of mixing all the datasets, it is common to divide training into multiple stages where each stage uses data from different domains to maximize the impact of the domain-specific training data.
}

\hl{
Two-stage fine-tuning~\citep{1604.02201} is a commonly used domain adaptation technique. Suppose there are $N$ datasets where $N-1$ of them are out-of-domain datasets and one of them is the in-domain dataset. MT model is trained on the mixture of $N$ datasets (or $N-1$ out-of-domain datasets) until convergence in the first stage and continuously trained on the in-domain dataset until convergence. Mixed fine-tuning~\cite{raj17} fine-tunes the mixture of $N$ datasets in the last step. \citep{imankulova2019exploiting} and \citep{dabre-etal-2019-exploiting} showed that training in multiple stages where each stage contains different proportions of various types of training data leads to the best results. Thus, in our case where there are more than two datasets, we modify the two-stage fine-tuning to multistage fine-tuning.
}

To make the transition between fine-tuning stages more smooth, we modify the two-stage fine-tuning to a multistage fine-tuning method. We first sort all the datasets into a list by the similarity with the in-domain dataset in terms of average per-token log-likelihood from the smallest to the largest given by the in-domain LM as described in \Sec{corpus_analysis}. For English--Japanese the sorted datasets are (A, T, C) and for English--Chinese the sorted datasets are (N, T, C).
There are $N$ stages, in the $i_{th}$ stage, the first $i$ datasets in the list are combined to train or fine-tune the model.

\subsection{Experimental Results}
\label{sec:experimental_results}
\noindent \textbf{Main Results.} We illustrate the effectiveness of the proposed Coursera corpus with the multistage fine-tuning domain adaptation method through the MT results in Tables~\ref{tab:selected_results} and \ref{tab:selected_results_zh_en} for English--Japanese and English--Chinese translations, respectively. We observed that: 

Firstly, leveraging the proposed in-domain dataset (\texttt{Mixed data}) yields huge performance improvements compared with using two out-of-domain datasets (\texttt{Mixed out-of-domain data}). For example, in the English\ra Japanese direction, the BLEU score improvement is $4.1$ points, and in the Chinese\ra English direction, the BLEU score improved from $21.6$ to $28.2$, achieving an improvement of $6.6$ points. Note that because our dataset itself is small, it does not work well on its own due to the overfitting problem.

Secondly, \texttt{Multistage fine-tuning} shows more robust if not better performance than \texttt{Mixed data}, \texttt{two-stage fine-tuning}~\citep{1604.02201} or \texttt{mixed fine-tuning}~\citep{raj17}. For English--Japanese MT experiments, multistage fine-tuning shows 0.1 and 0.4 BLEU score improvements compared with previous methods on Ja\ra En and En\ra Ja directions, respectively. For English--Chinese MT experiments, multistage fine-tuning also achieved the best results in both translation directions.

\noindent \textbf{More detailed results} are shown in~\Tab{results}, which summarizes the BLEU scores of all the MT systems trained up to five training stages for the Japanese--English tasks.
The training schedule with all the stages, i.e., A{\ra}AT{\ra}ATC{\ra}
TC{\ra}C (A5) did not achieve the best results in either translation direction. For the \task{Ja}{En} task, an intermediate model, A{\ra}AT{\ra}ATC (A3), achieved the best BLEU score with a gain of more than 20 points over the model trained only on the in-domain parallel data (B16).  For the reverse direction, that is, the \task{En}{Ja} task, the schedule A{\ra}AT{\ra}ATC{\ra}C (A6) achieved the best BLEU score with a 11.1 point BLEU gain.
Whenever additional training data were introduced (marked \colorbox{cR}{red} in \Tab{results}), the BLEU scores were improved.\footnote{Compare the pairs (A1, A2), (A2, A3), (A1, A10), (B2, B3), (C3, C4), (C10, C11), (D3, D4), (D10, D11), and (A1, E14).}
This is mostly in line with the observations of \citep{raj17}; starting from out-of-domain data only and ending with a mixture of out-of- and in-domain data gives the best results for in-domain translation. As shown in \Tab{datasetsimilarity}, A is more dissimilar to C, and T more similar to C. As such, it seems reasonable to gradually introduce the in-domain data by relying on related domain data for intermediate training steps.
According to~\citep{dabre-etal-2019-exploiting}, the final stage of fine-tuning on C should provide the best translation quality.  However, in our setting, this holds true only for some cases in the \task{En}{Ja} task, suggesting the necessity of hyper-parameter tuning for fine-tuning on deflated training data (marked \colorbox{cB}{blue} in \Tab{results}). In contrast, training a model directly in one stage on ATC (B10) yielded significantly lower results than the multistage results.

\section{Conclusion}
\label{sec:conclusion}
In this paper, we proposed a framework to create corpora for the lecture transcript translation.  
\hl{We proposed a parallel sentence extraction algorithm that extracts parallel sentences from parallel lecture transcripts, which is a DP algorithm based on machine translation and cosine similarity over sentence embeddings. Experimental results show a higher $F_1$ score compared with using BLEU-score similarity measurement or using the RoBERTa, LASER, or sentBERT embeddings.}

With the sentence extraction algorithm, we created an English--Japanese Coursera corpus with 50,543, 555, and 2,068 lines of train, development, and test sets, respectively. We also created an English--Chinese Coursera corpus with 40,074, 865, and 2,009 lines of train, development, and test sets, respectively. The quality of the test and development sets is ensured through manual filtering so that they can be used to reliably benchmark translation performance and tune MT models.

\hl{We illustrated the effectiveness of the proposed Coursera corpora through MT experiments. For English--Japanese and English--Chinese lecture translation, leveraging Coursera corpora brings large translation performance improvement measured by BLEU scores. Additionally, we modified the two-stage fine-tuning algorithm for multiple datasets and called it multistage fine-tuning. Experimental results showed more robust if not better performance compared with previous domain adaptation methods.}

Our future work aims to explore the method of multilingualism by constructing a multilingual corpus with a multilingual sentence alignment algorithm. To maximize the use of the corpora, we focus on adaptation techniques for low-resource domains.

\section{Acknowledgment}
Part of this work was conducted under the program ``Research and Development of Enhanced Multilingual and Multipurpose Speech Translation System'' of the Ministry of Internal Affairs and Communications (MIC), Japan.
This work was also supported by JSPS KAKENHI Grant Number JP23H03454 and JSPS Research Fellow for Young Scientists (DC1).

\bibliography{reference}
\end{document}